\documentclass[conference]{IEEEtran}
\IEEEoverridecommandlockouts
\usepackage{amsmath,amsfonts}
\usepackage[noend]{algorithmic}
\usepackage{algorithm}
\usepackage{array}
\usepackage[caption=false,font=normalsize,labelfont=sf,textfont=sf]{subfig}
\usepackage{textcomp}
\usepackage{stfloats}
\usepackage{url}
\usepackage{verbatim}
\usepackage{graphicx}
\usepackage{cite}
\usepackage{xcolor} 
\usepackage{booktabs} 
\usepackage{longtable}
\usepackage[linewidth=1pt]{mdframed}
\usepackage{booktabs} 
\usepackage{multirow}
\usepackage{graphicx}
\usepackage[colorlinks=true,linkcolor=black]{hyperref}

\begin{document}
\bibliographystyle{unsrt} 
\title{Lightweight Neural Path Planning
}

\author{Jinsong Li, Shaochen Wang, Ziyang Chen, Zhen Kan, and Jun Yu 
\thanks{This work was supported in part by the National Natural Science Foundation of China under Grant 62173314 and U2013601.}
\thanks{J. Li, S. Wang, Z. Chen, Z. Kan, and J. Yu (corresponding author) are with the Department of Automation, University of Science and Technology of China, Hefei, China.}
}

\markboth{Journal of \LaTeX\ Class Files,~Vol.~14, No.~8, August~2021}%
{Shell \MakeLowercase{\textit{et al.}}: A Sample Article Using IEEEtran.cls for IEEE Journals}


\maketitle


\begin{abstract}
Learning-based path planning is becoming a promising robot navigation methodology due to its adaptability to various environments. However, the expensive computing and storage associated with networks impose significant challenges for their deployment on low-cost robots. Motivated by this practical challenge, we develop a lightweight neural path planning architecture with a dual input network and a hybrid sampler for resource-constrained robotic systems. Our architecture is designed with efficient task feature extraction and fusion modules to translate the given planning instance into a guidance map. The hybrid sampler is then applied to restrict the planning within the prospective regions indicated by the guide map. To enable the network training, we further construct a publicly available dataset with various successful planning instances. Numerical simulations and physical experiments demonstrate that, compared with baseline approaches, our approach has nearly an order of magnitude fewer model size and five times lower computational while achieving promising performance. Besides, our approach can also accelerate the planning convergence process with fewer planning iterations compared to sample-based methods. The code and the dataset are available at 
\url{https://github.com/LeeXiaosong/Lightweight-planning}
\end{abstract}


\section{Introduction}
Planning a collision-free path from an initial location to the destination in complex environments is a long standing problem \cite{pathplanning},
and also the basic element in applications, such as autonomous driving \cite{7339478} and human–robot collaborations \cite{9568764}.
Classical path planning methods include potential field based methods \cite{PFM}, graph-based methods (e.g., Dijkstra \cite{Dijkstra} and A* algorithm \cite{Astar}), and sampling-based methods (e.g., probabilistic roadmap method (PRM) \cite{PRM}, rapidly exploring random tree (RRT) \cite{RRT}, RRT* \cite{RRTstar}, and their variants). However, potential field based methods can suffer from local minimum and graph-based methods are generally computationally expensive due to the abstraction of the environment into a grid world and thus can only be used for low-dimensional spaces. Due to the probabilistic completeness and scalability, sampling-based methods have been widely applied for the path planning of mobile robots, but they converge slowly to a feasible (optimal if RRT* is used) path, since many unnecessary samples are drawn from regions that optimal paths are not likely to exist. Recently, learning-based methods that leverage deep neural networks to encode the environment have achieved promising results in path planning. However, most existing learning-based methods are resource demanding and require enormous computing resources and storage space. As resource-constrained robotic systems are more involved in our daily life (e.g., home service and healthcare), the low power consumption, as well as limited computation and storage, are imposing significant challenges for learning-based path planning approaches. Intensive computing resources and real-time requirements become the barriers for the implementation of deep-learning-based technologies in robotic systems. Therefore, this work is particularly motivated to design a learning-based path planner for resource-constrained robotic systems with good performance but low computational footprint, real-time implementation, and lightweight model size.

\begin{figure}[t]
\begin{center}
\begin{minipage}[t]{0.45\linewidth}
    \centering
    \includegraphics[width=\textwidth]{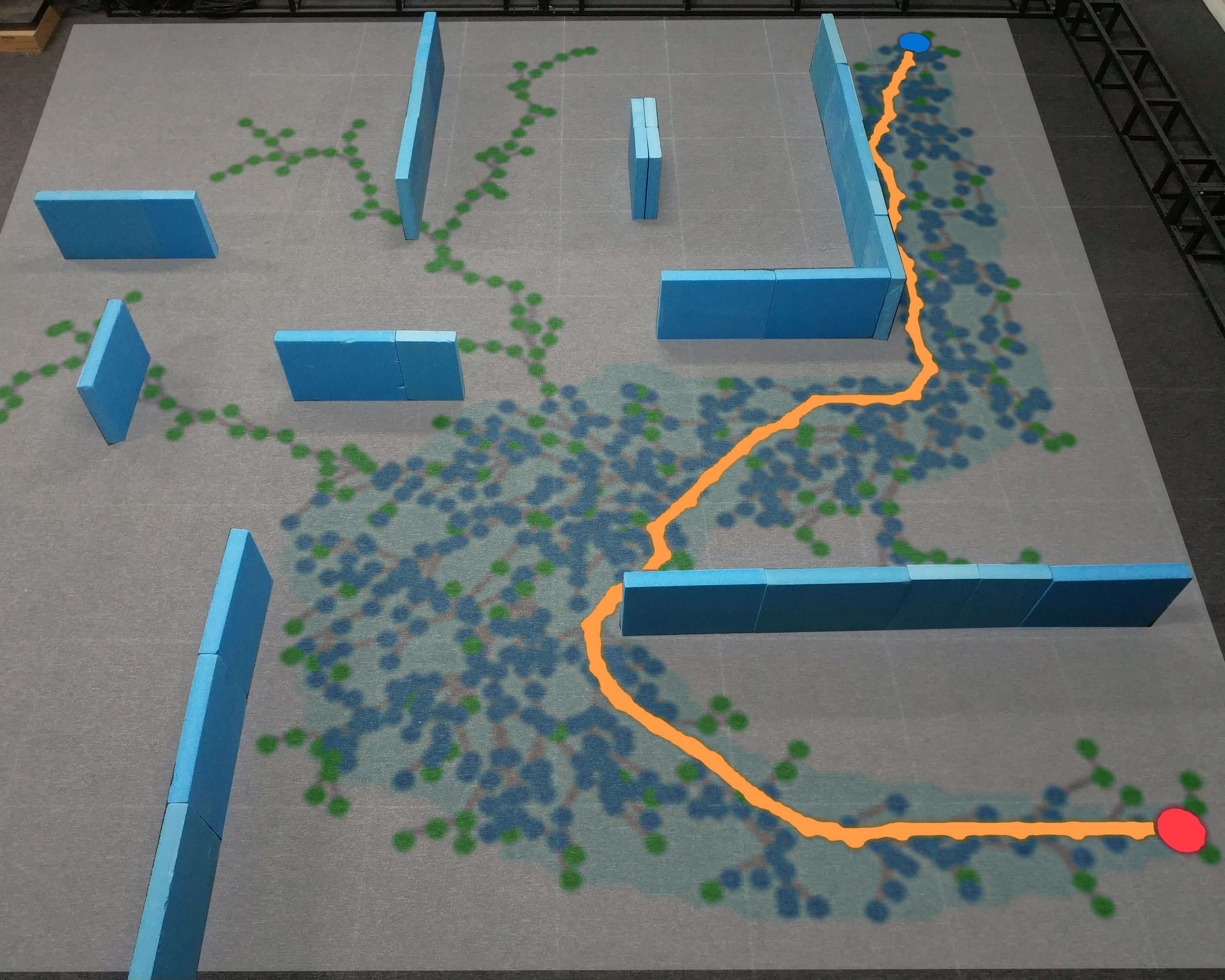}
    \centerline{\footnotesize{(a) Ours}}
\end{minipage}%
\quad
\begin{minipage}[t]{0.45\linewidth}
    \centering
    \includegraphics[width=\textwidth]{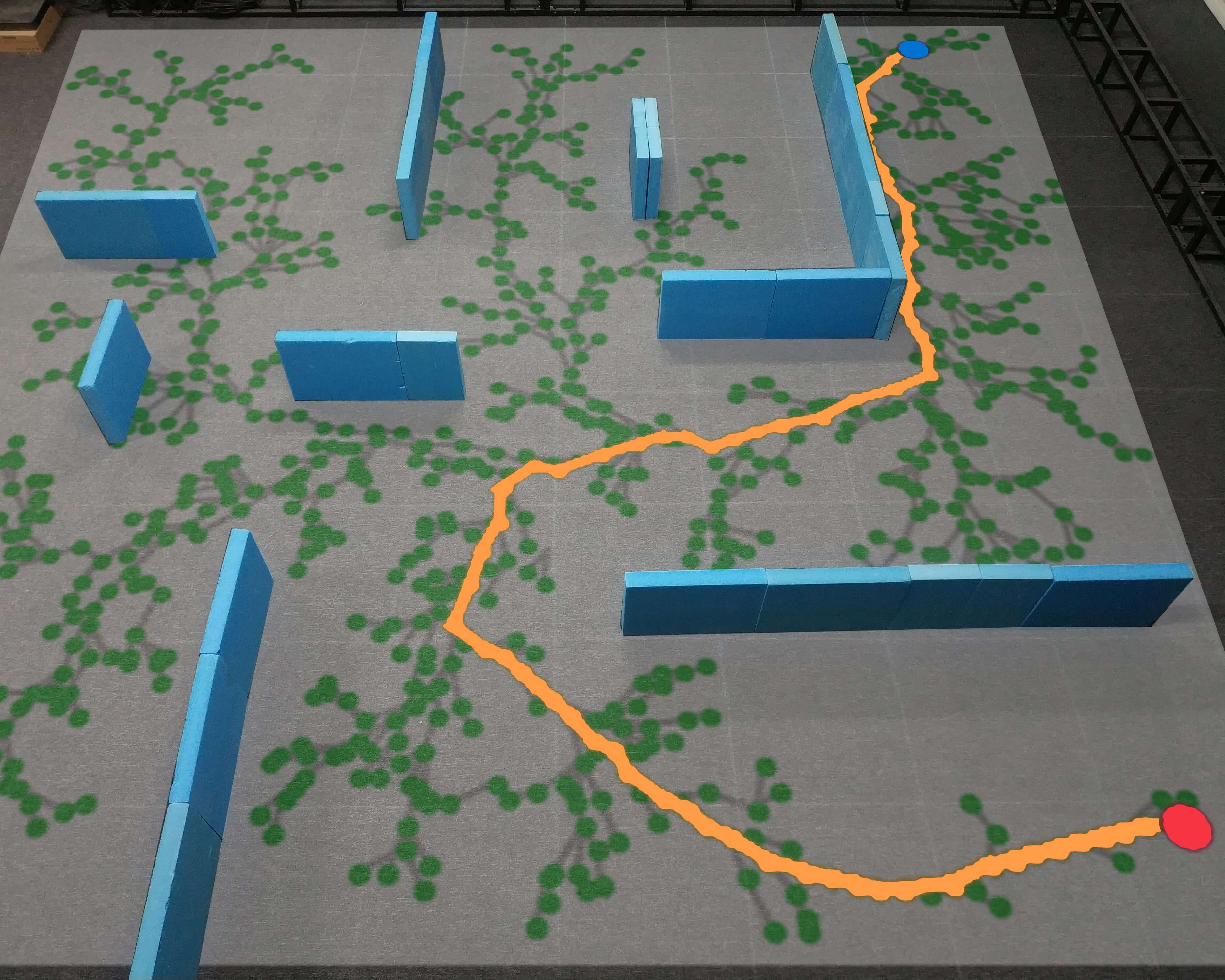}
    \centerline{\footnotesize{(b) RRT*}}
\end{minipage}
\end{center}
\caption{Path planning using our lightweight learning-based algorithm and traditional RRT* on mobile robot platforms.
}  
\label{realworld}
\end{figure}

Learning-based path planning is a recent research focus \cite{LearningM}. The works of \cite{RLplanning1} and \cite{RLplanning2} leverage reinforcement learning (RL) to model the interactions between the robot and environment to address real-time path planning problems. RL is also integrated with temporal logic specifications in \cite{LTL1} and \cite{LTL2} for motion planning of complex tasks. Since RL \cite{9679819}  suffers from low sampling efficiency, especially for long learning processes, supervised learning is exploited to guide the planning processes or directly generate viable paths. For instance, OracleNet \cite{OracleNet} encodes the trajectory history and predicts the next location at each step by learning from expert trajectories. Convolutional neural networks (CNNs) are also widely adopted for path planning. In \cite{NeuralA}, a novel data-driven search method was developed by designing a convolutional encoder to form an end-to-end trainable neural network for path planning problems. In \cite{CVAE}, the conditional variational autoencoder (CVAE) is utilized to generate potential distributions of promising states from demonstrations and used as a module in RRT* for optimal path planning. In \cite{cGAN}, generative adversarial network (GAN) was exploited to facilitate the prediction of promising regions used in non-uniform sampling.  The work of \cite{cGAN} is then extended in \cite{GAN1} and \cite{GAN2} for path planning in 3-D spaces with improved connectivity of promising regions. Despite recent progress of guiding the sampling of RRT/RRT* via deep neural networks (e.g., CNNs, GAN, etc.), these approaches generally require considerable computing resources and storage space and thus are hard to be deployed on resource-constrained robotic systems.

In this work, we develop a lightweight learning-based path planning algorithm for resource-constrained robotic systems with limited computation and storage capabilities. Specifically, to facilitate the training of neural networks, we first construct a rich dataset containing 80, 000 planing scenarios including mazes, corridors, chambers, junctions, and common objects (e.g., columns, triangles, squares). Each scenario contains expert trajectories generated by RRT as the ground truth. By trading off the network size, inference time, and planning performance, we then develop a lightweight neural path planning architecture based on a fast CNN generator and a hybrid state sampler for improved sampling efficiency. As highlighted in Fig. \ref{realworld}, our approach can focus on promising regions while conventional RRT has to sample over the entire workspace. In addition, the lightweight design enables the deployment on resource-constrained robotic systems and can be flexibly integrated with other sampling-based approaches to accelerate path planning. 

The main contributions can be summarized as follows:
\begin{itemize}
	\item Construct a public dataset with abundant successful path planning examples, which can be used as training scenarios for learning-based methods or verification examples.    
	\item Develop a lightweight neural path planner for resource-constrained robotic systems.
	\item Demonstrate its efficiency over SOTA methods and validate its effectiveness in physical experiments.
\end{itemize}

\section{Problem Formulation}
Let $\mathcal{X}$ represent a $d$-dimensional configuration space of the mobile robot $\mathcal{A}$. The obstacles are denoted by  $\mathcal{X}_{obs}$ and the obstacle-free space is denoted by $\mathcal{X}_{free}=\mathcal{X} \backslash \mathcal{X}_{obs}$.
The initial state and goal state of the robot are denoted by $x_{init}\in \mathcal{X}$ and $\mathcal{X}_{goal}\subset \mathcal{X}_{free}$, respectively.
A continuous path $\sigma:[0, 1]\to \mathbb{R}^{d}$ is called feasible if $\sigma(0)\subset \mathcal{X}_{init}$, $\sigma(1)\subset \mathcal{X}_{goal}$, and $\sigma(\tau)\in \mathcal{X}_{free}$ for all $\tau\in [0,1]$. Let $\Phi=\{\sigma_{i}\}_{i=1}^{n}$ be the set of feasible paths. Given the tuple $(\mathcal{X},x_{init},\mathcal{X}_{goal})$ and a
cost function $\upsilon:\Phi\to\mathbb{R}_{\geq0}$, the goal of
this work is to find an optimal path $\sigma^{*}$ such that $\sigma^{*}=\arg\min_{\sigma\in\Phi}\upsilon(\sigma)$. In this work, we define the cost $\upsilon(\sigma)$ as the geometric Euclidean distance of the path $\sigma\in\Phi$.

\section{Approach}
\begin{figure*}[htbp]
\centering
\includegraphics[width=1.95\columnwidth,height=0.21\linewidth]{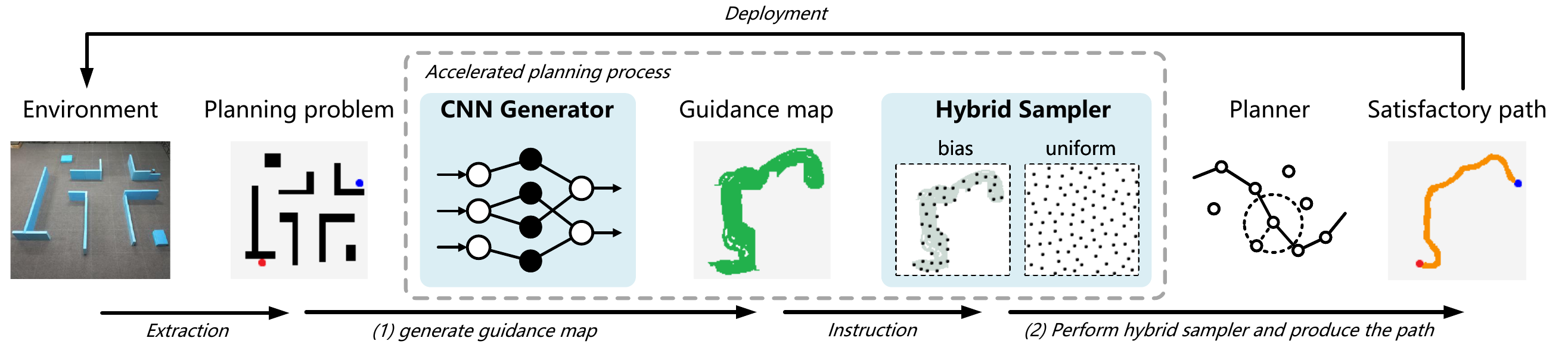}
\caption{The Schematic diagram of our lightweight neural path planning  architecture. (1) The generator takes the extracted planning map as input and outputs a guidance map.
(2) The sampled states are concentrated in the promising regions by the hybrid sampler.}
\label{liucheng}
\end{figure*}

As illustrated in Fig. \ref{liucheng}, 
the proposed lightweight path planning method consists of two main components: 1) a CNN generator that outputs a guidance map and 2) a hybrid sampler that selects the state in the promising region with higher probability. 

\subsection{Generator Network}
In learning-based methods, deep neural networks are leveraged to accelerate the sampling process to allow for faster path planning. Hence, a CNN generator is developed, which learns the sampling distribution from successful planning experience and predicts a prospective sampling region $\mathcal{M}_g$ (i.e., the guidance map) from the start point to the target point by feeding the task instance $(\mathcal{X}, x_{init}, \mathcal{X}_{goal})$ into network. To achieve a lightweight design, the guidance maps $\mathcal{M}_g$ is generated based on the pix2pix method \cite{pix2pix} 
that includes 
a lightweight generator model with a dual input branch, a PatchGAN discriminator, and a joint objective function.

\begin{figure*}[htbp]
\centering
\includegraphics[width=1.85\columnwidth,height=0.43\linewidth]{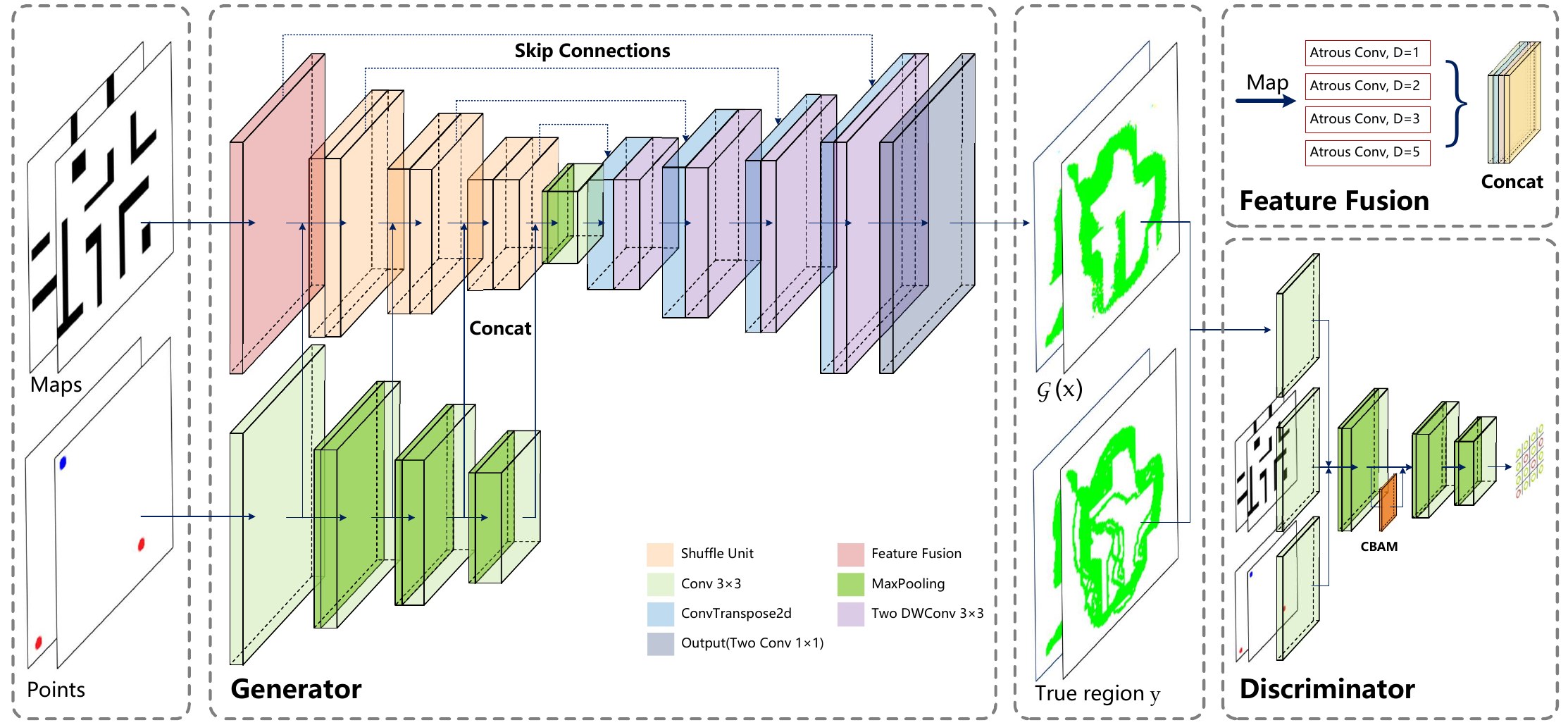}
\caption{Overall architecture of the proposed guidance map generator.}
\label{net}
\end{figure*}

As illustrated by the framework of the CNN generator in Fig. \ref{net}, the pair $x = (m,p)$ composed of the environment map $m\in \mathcal{M}_e$ and the initial and target points $p\in \mathcal{P}$ 
are taken as input to the generator.
The UNet-like structure is leveraged to design the generator network 
by combining a lightweight feature extracting module 
and a novel convolutional module to obtain long contextual information. 
The generated regions $\mathcal{M}_g=\mathcal{G}(x)$ will then be utilized via biased sampling to enable efficient path planning. 

\textbf{Lightweight design:}
An ordinary convolution kernel $(D_k \times D_k \times n)$  applied to an image with size $H \times W \times N$ 
usually takes $D_k \times D_k \times N \times n$ parameters.
As for depthwise separable convolution (DWConv) \cite{DSConv}, only $D_k \times D_k \times N + N \times n$ parameters are needed to produce the same size, which is
$1/(1/N+1/{D_k}^2) \approx {D_k}^2$ of ordinary convolution.
Thus, we propose a lightweight backbone modified from the basic UNet\cite{Unet}
for the trade-off between model size and accuracy  . 

\begin{figure}[htbp]
\begin{center}
\begin{minipage}[t]{0.38\linewidth}
    \centering
    \includegraphics[width=\textwidth]{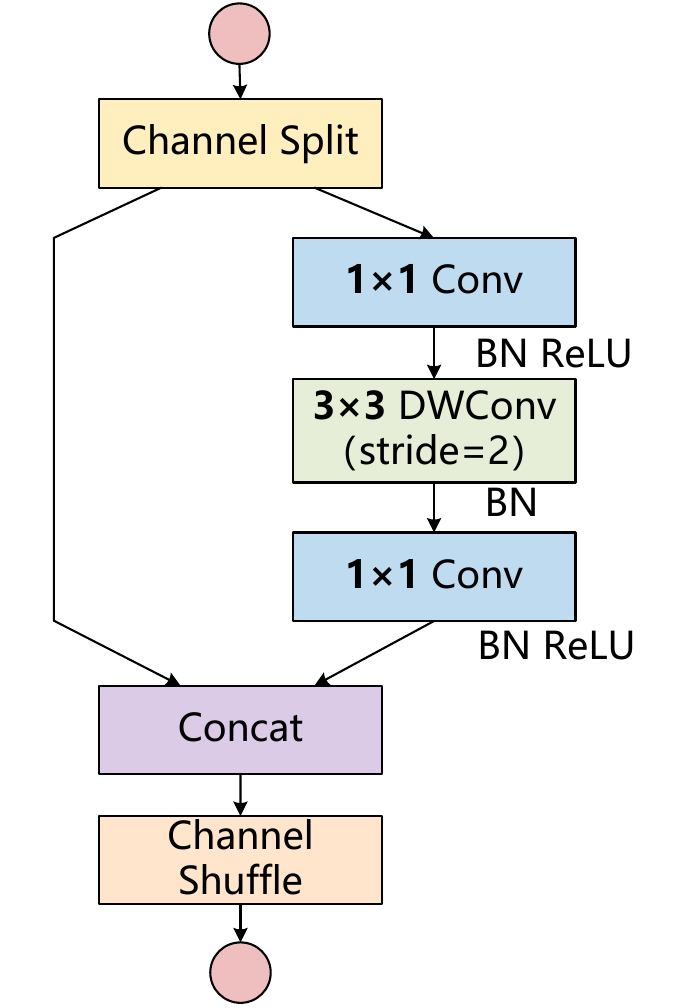}
    \centerline{\footnotesize{(a)}}
\end{minipage}%
\quad
\quad
\begin{minipage}[t]{0.46\linewidth}
    \centering
    \includegraphics[width=\textwidth]{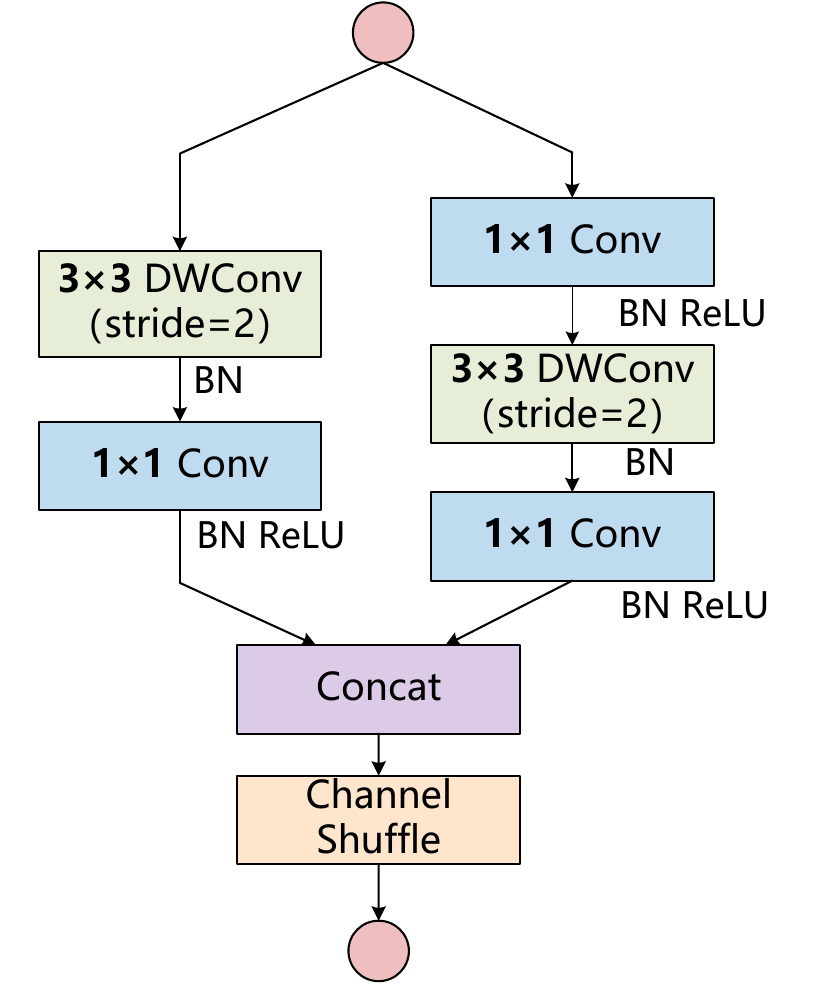}
    \centerline{\footnotesize{(b)}}
\end{minipage}
\end{center}
\caption{The feature extractor used in our model. (a) The basic block. (b) The downsampling block}
\label{shuffle}
\end{figure}
Instead of using $3\times 3$ ordinary convolution, 
our model makes use of a shuffle unit consisting of DWConv and $1\times 1$ Conv 
as feature extractors in the main encoding path.
As shown in Fig. \ref{shuffle}, the input feature is split into two branches. Half of the channels goes through the $1\times 1$ Conv layer, the $3\times 3$ DWConv layer, and the $1\times 1$ Conv layer in turn,
while the rest channels are unchanged in the basic block as shown in Fig. \ref{shuffle}(a).
In the downsampling block in Fig. \ref{shuffle}(b), two successive operations 
including $1\times 1$ Conv layer and $3\times 3$ DWConv layer  
are performed on the input features, respectively. The channel shuffle operation is then executed after concatenating these branches.
In the decoding path, the skip connection is employed, 
and we stack two $3\times 3$ DWConv and ConvTranspose2d in an interlaced way as an up-sampling layer.
Thus, the computational cost of the overall network is lowered while maintaining the output quality, due to the shuffle unit and the lower parameters of DWConv.

\textbf{Multi-scale feature fusion:}
Lightweight techniques can reduce parameters, but it may lead to incomplete extraction of some obstacle features in distant regions in the encoding phase, resulting in degraded quality of predicted guidance regions. A common approach to address this issue is to increase the receptive field size of networks by stacking ordinary convolution layers or adopting larger convolution kernels; however it will lead to larger model size and more computational budge, limiting its applicability in resource-constrained robots. In order to reduce the model size without sacrificing planning accuracy, we design a feature fusion module to effectively capture long-range information. Since the atrous convolution can increase the convolution area without using an excessive amount of parameters, motivated by the works of DeepLab family \cite{DeepLab}, a fusion module is introduced behind the input interface of environment maps that uses four concurrent atrous convolutions with $3\times 3$ kernel but different dilation rates ($rate = 1,2,3,5$). The resulting features from four branches will be concatenated to a multi-scale feature map before subsequent processing.

\textbf{Dual input branch:}
To guarantee the matching between the task and generated area,
an additional path is used to extract the task information, e.g., the start and end points in the task map.
In the generator block of Fig. \ref{net}, the branch is divided into four stages,
each of which consists of an ordinary convolution followed by a max pooling operation.
After the convolution process, the task feature is delivered to the corresponding stage of the backbone 
and concatenated with the environment map feature.
The purpose of this structure is to allocate as few computational resources to encode task points features 
while ensuring the generator's performance, instead of simply concatenating or summing these two maps.

\subsection{Discriminator Network}
When training the generator $\mathcal{G}$, 
an efficient discriminator $\mathcal{D}$ modified from Patch-GAN is employed, 
since it can produce a $N \times N$ probability matrix instead of a single $0/1$ scalar at once.
In Fig. \ref{net}, 
the environment map and task points, 
as well as the generated $\mathcal{G}(x)$ or the ground truth $y$, 
are fed into a four-layer convolutional network, 
where each layer consists of Conv+Norm+ReLU block and max pooling operation.
And it finally outputs a matrix of $1$ or $0$, indicating whether each patch is real or fake.
Furthermore, 
an attention module CBAM \cite{atCBAM} is introduced to the second layer following the basic block 
to help focus on important features.


\subsection{Training objective} 
The network is desired to not only generate prospective planning regions
but also can be generalized to new environments that have not been trained.
Thus, a joint objective function is constructed in this section. 

\textbf{Adversarial loss:} The generator $\mathcal{G}$ is trained to minimize the objective function, while the discriminator $\mathcal{D}$ is trained to maximize it. Hence, 
the loss function is designed as
\begin{equation}\label{cganloss}
\begin{aligned}
\mathcal{L}_{cGAN}(\mathcal{G},\mathcal{D})&=\mathbb{E}_{x,y}[\log \mathcal{D}(x,y)] \\
&+ \mathbb{E}_{x,z}[\log(1- \mathcal{D}(x,\mathcal{G}(x,z)))].   
\end{aligned}
\end{equation}
In (\ref{cganloss}), to learn a  mapping from the random noise $z$ to the target image $y$ conditioned on $x = (m,p)$,
$\mathcal{G}$ aims at minimizing $\mathbb{E}_{x,z}[\log(1- \mathcal{D}(x,\mathcal{G}(x,z)))]$ through producing real-looking images.
In turn, $\mathcal{D}$ evaluates input images as fake/real to maximize $\mathbb{E}_{x,y}[\log \mathcal{D}(x,y)]$.

To avoid unstable training and the global distortion with free low frequency patterns, the
$L_1$ loss function is defined as
\begin{equation}\label{l1loss}
\mathcal{L}_{L_{1}}(\mathcal{G})=\mathbb{E}(\| y-\mathcal{G}(x) \|_{1} ),
\end{equation}
which is optimized during the training process to constrain the proximity of generated areas $\mathcal{G}(x)$ and the planned paths $y$ (also referred to as ground truth).
Since the boundary of path areas is matched by calculating the $L_{1}$ distance, 
the generator is capable of capturing the whole appearance of robot planning regions more correctly,
which is suitable for area reconstruction of the guidance map $\mathcal{M}_g$.

\textbf{Auxiliary loss:} 
The dominating term bias in the loss function may be caused by 
the imbalance of positive and negative samples of guidance maps, 
such as smaller predicted areas and bigger blank backgrounds.
To alleviate this,
the Dice Score Coefficient (DSC)\cite{DSCloss} is employed as
\begin{equation}
\mathcal{L}_{DSC}(\mathcal{G})=1- \frac{2(1-\hat{p})\hat{p}\cdot y+\gamma}{(1-\hat{p})\hat{p}\cdot y+\gamma}, 
\label{dscloss}
\end{equation}
where $\hat{p}=\mathcal{G}(x)$ denotes the predicted output and $y$ is the ground truth.
The parameter $\gamma$ is added to prevent (\ref{dscloss}) from undefined cases such as $y = \hat{p} = 0$.

Based on (\ref{cganloss}), (\ref{l1loss}), and (\ref{dscloss}), the final objective function is written as 
\begin{equation}\label{obloss}
\mathcal{L}(\mathcal{G},\mathcal{D})=\lambda_1 \mathcal{L}_{cGAN}(\mathcal{G},\mathcal{D}) + \lambda_2 \mathcal{L}_{L_{1}}(\mathcal{G}) + \lambda_3 \mathcal{L}_{DSC}(\mathcal{G}) 
\end{equation}
where the weighting factors $\lambda_1$, $\lambda_2$, $\lambda_3$ indicate relative importance.
The generator $G^{*}$ is obtained by solving the following optimization problem
\begin{equation}
\mathcal{G}^{*} = \arg \mathop{\min}_{\mathcal{G}} \mathop{\max}_{\mathcal{D}} \mathcal{L}(\mathcal{G},\mathcal{D}).
\end{equation}

\subsection{Hybrid Sampler}
Inspired by the work of \cite{cGAN}, a hybrid sampler is developed. As outlined in Algorithm 1, sampling from the guidance map $\mathcal{M}_g$ forces the planner to select the state $x_{rand}$ in the region connecting the initial and target points. Besides sampling from $\mathcal{M}_g$, states can also be uniformly sampled from the environment, though redundant states may be added. The BiasFactor is a tuning hyper-parameter that balances the efficiency and exploration via flexible switching between biased sampling and uniform sampling, while ensuring the probabilistic completeness. 

\begin{algorithm}[htb]
\algsetup{linenosize=\scriptsize} \small
\caption{: Hybrid Sampler}\label{BSampling}  
\hspace*{0.02in}{\bf Input:} environment map $\mathcal{M}_e$, guidance map $\mathcal{M}_g$ \\
\hspace*{0.02in}{\bf Output:} state $x_{rand}$
\begin{algorithmic}[1]
\IF {$Rand()<BiasFactor$}
    \STATE $x_{rand}\leftarrow RandomSample(\mathcal{M}_e)$
\ELSE
    \STATE $x_{rand}\leftarrow RandomSample(\mathcal{M}_g)$
\ENDIF
\RETURN {$x_{rand}$}
\end{algorithmic}
\end{algorithm}

\section{Results}
Numerical simulations and physical experiments are carried out in this section to demonstrate the developed learning-based path planning algorithm. 

\subsection{Datasets Construction}

Due to the lack of high-quality datasets of environment maps with successful planning experience for deep learning techniques, we construct a dataset consisting of environment maps and points maps with expert knowledge. Table \ref{dataTB} gives a brief presentation of the dataset, which includes $80,000$ scenarios abstracted from tough environmental features including mazes, corridors, chambers, junctions, to other common objects (e.g., columns, triangles, squares). The maps are in RGB format with the obstacle-free space $\mathcal{X}_{free}$ and obstacle regions $\mathcal{X}_{obs}$ represented in white and black, respectively. For each map in the dataset, $RRT$ was executed $50$ times with randomly selected start and end points (i.e., blue and red point in the map, respectively). The generated feasible paths are stacked together to form the guidance map (i.e., the green area in the map). The above process was performed on the whole dataset, producing $68,000$ training and $12,000$ test set. The dataset has been released\footnote{The dataset is publicly availible at: \url{https://drive.google.com/file/d/1\_XBhPdxuhU\_pVKeXqC3br3T9Wcw-MnVt/view}}.

\begin{table}[htbp]
\caption{A brief exhibition of our dataset} 
\centering
\fontsize{7}{10}\selectfont    
\setlength{\tabcolsep}{7pt} 
\renewcommand{\arraystretch}{0.05} 
\begin{tabular}{ccccc|c}
\toprule 
Scene~ & ~Map~  &  ~Point  & ~Task  & Guidance &  Overview \\ 
\midrule 
Maze &
\begin{minipage}[b]{0.1\columnwidth}
\raisebox{-.4\height}{\setlength{\fboxsep}{0pt}\fbox{\includegraphics[width=\linewidth]{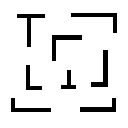}}}
\end{minipage} &
\begin{minipage}[b]{0.1\columnwidth}
\raisebox{-.4\height}{\setlength{\fboxsep}{0pt}\fbox{\includegraphics[width=\linewidth]{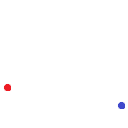}}}
\end{minipage} & 
\begin{minipage}[b]{0.1\columnwidth}
\raisebox{-.4\height}{\setlength{\fboxsep}{0pt}\fbox{\includegraphics[width=\linewidth]{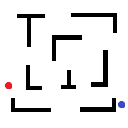}}}
\end{minipage} & 
\begin{minipage}[b]{0.1\columnwidth}
\raisebox{-.4\height}{\setlength{\fboxsep}{0pt}\fbox{\includegraphics[width=\linewidth]{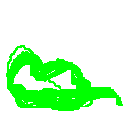}}}
\end{minipage} &
\begin{minipage}[b]{0.1\columnwidth}
\raisebox{-.4\height}{\setlength{\fboxsep}{0pt}\fbox{\includegraphics[width=\linewidth]{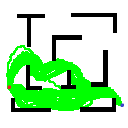}}}
\end{minipage}     
\\
Corridor &
\begin{minipage}[b]{0.1\columnwidth}
\raisebox{-.4\height}{\setlength{\fboxsep}{0pt}\fbox{\includegraphics[width=\linewidth]{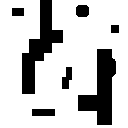}}}
\end{minipage} &
\begin{minipage}[b]{0.1\columnwidth}
\raisebox{-.4\height}{\setlength{\fboxsep}{0pt}\fbox{\includegraphics[width=\linewidth]{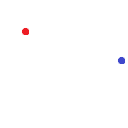}}}
\end{minipage} & 
\begin{minipage}[b]{0.1\columnwidth}
\raisebox{-.4\height}{\setlength{\fboxsep}{0pt}\fbox{\includegraphics[width=\linewidth]{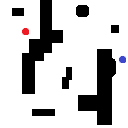}}}
\end{minipage} & 
\begin{minipage}[b]{0.1\columnwidth}
\raisebox{-.4\height}{\setlength{\fboxsep}{0pt}\fbox{\includegraphics[width=\linewidth]{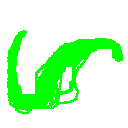}}}
\end{minipage} &  
\begin{minipage}[b]{0.1\columnwidth}
\raisebox{-.4\height}{\setlength{\fboxsep}{0pt}\fbox{\includegraphics[width=\linewidth]{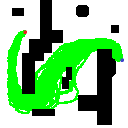}}}
\end{minipage}  
\\
Rooms &
\begin{minipage}[b]{0.1\columnwidth}
\raisebox{-.4\height}{\setlength{\fboxsep}{0pt}\fbox{\includegraphics[width=\linewidth]{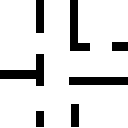}}}
\end{minipage} &
\begin{minipage}[b]{0.1\columnwidth}
\raisebox{-.4\height}{\setlength{\fboxsep}{0pt}\fbox{\includegraphics[width=\linewidth]{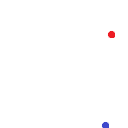}}}
\end{minipage} & 
\begin{minipage}[b]{0.1\columnwidth}
\raisebox{-.4\height}{\setlength{\fboxsep}{0pt}\fbox{\includegraphics[width=\linewidth]{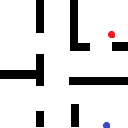}}}
\end{minipage} & 
\begin{minipage}[b]{0.1\columnwidth}
\raisebox{-.4\height}{\setlength{\fboxsep}{0pt}\fbox{\includegraphics[width=\linewidth]{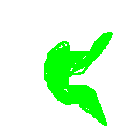}}}
\end{minipage} &  
\begin{minipage}[b]{0.1\columnwidth}
\raisebox{-.4\height}{\setlength{\fboxsep}{0pt}\fbox{\includegraphics[width=\linewidth]{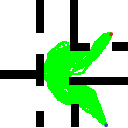}}}
\end{minipage}   
\\
Junction &
\begin{minipage}[b]{0.1\columnwidth}
\raisebox{-.4\height}{\setlength{\fboxsep}{0pt}\fbox{\includegraphics[width=\linewidth]{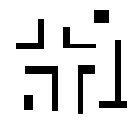}}}
\end{minipage} &
\begin{minipage}[b]{0.1\columnwidth}
\raisebox{-.4\height}{\setlength{\fboxsep}{0pt}\fbox{\includegraphics[width=\linewidth]{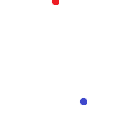}}}
\end{minipage} & 
\begin{minipage}[b]{0.1\columnwidth}
\raisebox{-.4\height}{\setlength{\fboxsep}{0pt}\fbox{\includegraphics[width=\linewidth]{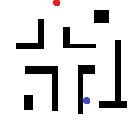}}}
\end{minipage} & 
\begin{minipage}[b]{0.1\columnwidth}
\raisebox{-.4\height}{\setlength{\fboxsep}{0pt}\fbox{\includegraphics[width=\linewidth]{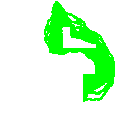}}}
\end{minipage} &  
\begin{minipage}[b]{0.1\columnwidth}
\raisebox{-.4\height}{\setlength{\fboxsep}{0pt}\fbox{\includegraphics[width=\linewidth]{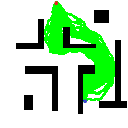}}}
\end{minipage}   
\\
Columns  &
\begin{minipage}[b]{0.1\columnwidth}
\raisebox{-.4\height}{\setlength{\fboxsep}{0pt}\fbox{\includegraphics[width=\linewidth]{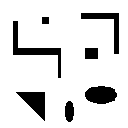}}}
\end{minipage} &
\begin{minipage}[b]{0.1\columnwidth}
\raisebox{-.4\height}{\setlength{\fboxsep}{0pt}\fbox{\includegraphics[width=\linewidth]{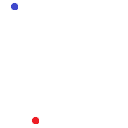}}}
\end{minipage} & 
\begin{minipage}[b]{0.1\columnwidth}
\raisebox{-.4\height}{\setlength{\fboxsep}{0pt}\fbox{\includegraphics[width=\linewidth]{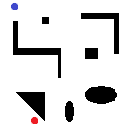}}}
\end{minipage} & 
\begin{minipage}[b]{0.1\columnwidth}
\raisebox{-.4\height}{\setlength{\fboxsep}{0pt}\fbox{\includegraphics[width=\linewidth]{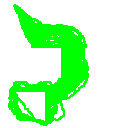}}}
\end{minipage} &  
\begin{minipage}[b]{0.1\columnwidth}
\raisebox{-.4\height}{\setlength{\fboxsep}{0pt}\fbox{\includegraphics[width=\linewidth]{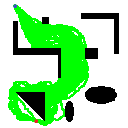}}}
\end{minipage}   
\\
\bottomrule 
\end{tabular}
\label{dataTB}
\end{table}

\subsection{Evaluation Metrics}
The following metrics are used to evaluate the model's effectiveness and improvement over sampling-based methods.

\textbf{Lightweight and accuracy metrics:}
To measure the performance of image generation, the mean Dice Coefficient (mDice) and the mean Intersection over Union (mIoU) are employed. Given the goal of lightweight design for resource-constrained robot platforms, the number of parameters, GFLOPs, and the inference time are also evaluated.

The following widely-used methods are selected as baselines.
\begin{enumerate}
\item 
Original UNet \cite{Unet}: A network with five layers of the contracting and expanding paths.
\item 
Res-UNet \cite{ResUnet}: Incorporate residual blocks into original UNet to prevent the network degradation. 
\item 
Nest-UNet \cite{NestUnet}: Redesign skip connections to enable flexible feature fusion.
\item 
Fast-FCN \cite{FFCN}: Use Joint Pyramid Upsampling (JPU) to reduce time and memory consuming.
\end{enumerate}

\textbf{Planning efficiency metrics:}
By introducing the hybrid sampler, the developed $Guided$-$RRT$ and $Guided$-$RRT^*$ are compared with original $RRT$ and $RRT^*$ to validate its performance of optimizing the path length and sampling cost.

\subsection{Implementation Details}
All numerical experiments are performed on a computer with Intel i9-10940X CPU and GeForce RTX 3090 GPU, which runs Ubuntu 18.04 and PyTorch version 3.7.  The images fed to the networks are cropped to $128 \times 128$ pixels and the size of input batch is $(N, C, H, W)$ before training. The generator and discriminator are trained using Adam optimizer with a learning rates of $0.005$ and $0.001$, respectively. The best model of each generator is selected after 25 training epochs. The BiasFactor is set to $0.9$ and the step-size of the planning algorithm is set to 2 throughout the path search phase.

\subsection{Evaluation Results}
\textbf{Comparison to CNNs models:}
Our method is compared with SOTA methods, such as Orig-Unet, Res-Unet, Nest-Unet, and Fast-FCN in this section. These baseline networks have the same feature scale and output layer made up of two $1 \times 1$ convolutions for consistency. Since these networks have a single encoding path, the environment map and task points are represented on a single image and fed to these networks. The best results in terms of mIou, mDice, GFLOPs, and Params are selected and listed in Table \ref{Acc}. 

\begin{table*}[htbp]
\caption{Comparison of different networks in mIou, mDice, GFLOPs, Params and inference time.} 
\label{FPS}
\centering
\fontsize{9}{10}\selectfont    
\setlength{\tabcolsep}{5pt} 
\renewcommand{\arraystretch}{0.8} 
\begin{tabular}{ccccccccccccccccc}
\toprule
\multirow{2}{*}{Network}  & & \multirow{2}{*}{mIoU} & & \multirow{2}{*}{mDice}  & & \multirow{2}{*}{GFLOPs} & & \multirow{2}{*}{Params}  & & \multicolumn{3}{c}{Inference time(in GPU)} & & \multicolumn{3}{c}{Inference time(in CPU)}  \\
\cmidrule[0.5pt]{11-13} 
\cmidrule[0.5pt]{15-17}
& & & & & & & & & & $128\times 128$ & &$256\times 256$ & & $128\times 128$ & &$256\times 256$  \\
\midrule 
Orig-Unet && 0.57 && 0.72 && 16.15 && 34.17 M && \textbf{4.0 ms} && 8.8 ms && 79.4 ms && 258.2 ms   \\
Res-Unet && 0.61 && 0.75 && 17.17 && 44.67 M && 4.1 ms && 9.5 ms && 73.1 ms && 257.9 ms   \\
Nest-UNet && 0.67 && 0.79 && 34.68 && 36.63 M && 5.4 ms && 18.5 ms && 118.6 ms && 476.7 ms   \\
Fast-FCN && 0.56 && 0.70 && 65.66 && 85.31 M && 13.1 ms && 16.9 ms && 80.2 ms && 262.0 ms   \\
\midrule 
Ours && \textbf{0.72} && \textbf{0.82} && \textbf{3.22} && \textbf{4.42 M} &&  4.5 ms && \textbf{5.7 ms} && \textbf{54.5 ms} && \textbf{192.0 ms}  \\
\bottomrule
\end{tabular}\vspace{0cm}
\label{Acc}
\end{table*}

Overall, our model outperforms these baselines in terms of the image similarity, i.e., $71.6\%$ mIoU and $82.2\%$ mDice.
Our model also shows substantial decrease in network size, i.e., only $4.42 M$ parameters and $3.22$ GFLOPs, enabling implementations on mobile robot platforms.
Nest-Unet gets the second-best score for image accuracy but it requires  heavy computing resources.
While feeding $128\times 128$ images to the network, 
a little gap exists between our model and UNet in the process time on GPU, 
which may be caused by the dual input design. 
However, 
our model shows the advantage of faster processing at $256 \times 256$ image resolution,
i.e., the time cost is decreased by $35.2\%$ over the fastest baseline method (UNet).

\begin{figure}[htbp]
\centering
\includegraphics[width=0.95\columnwidth,height=0.56\linewidth]{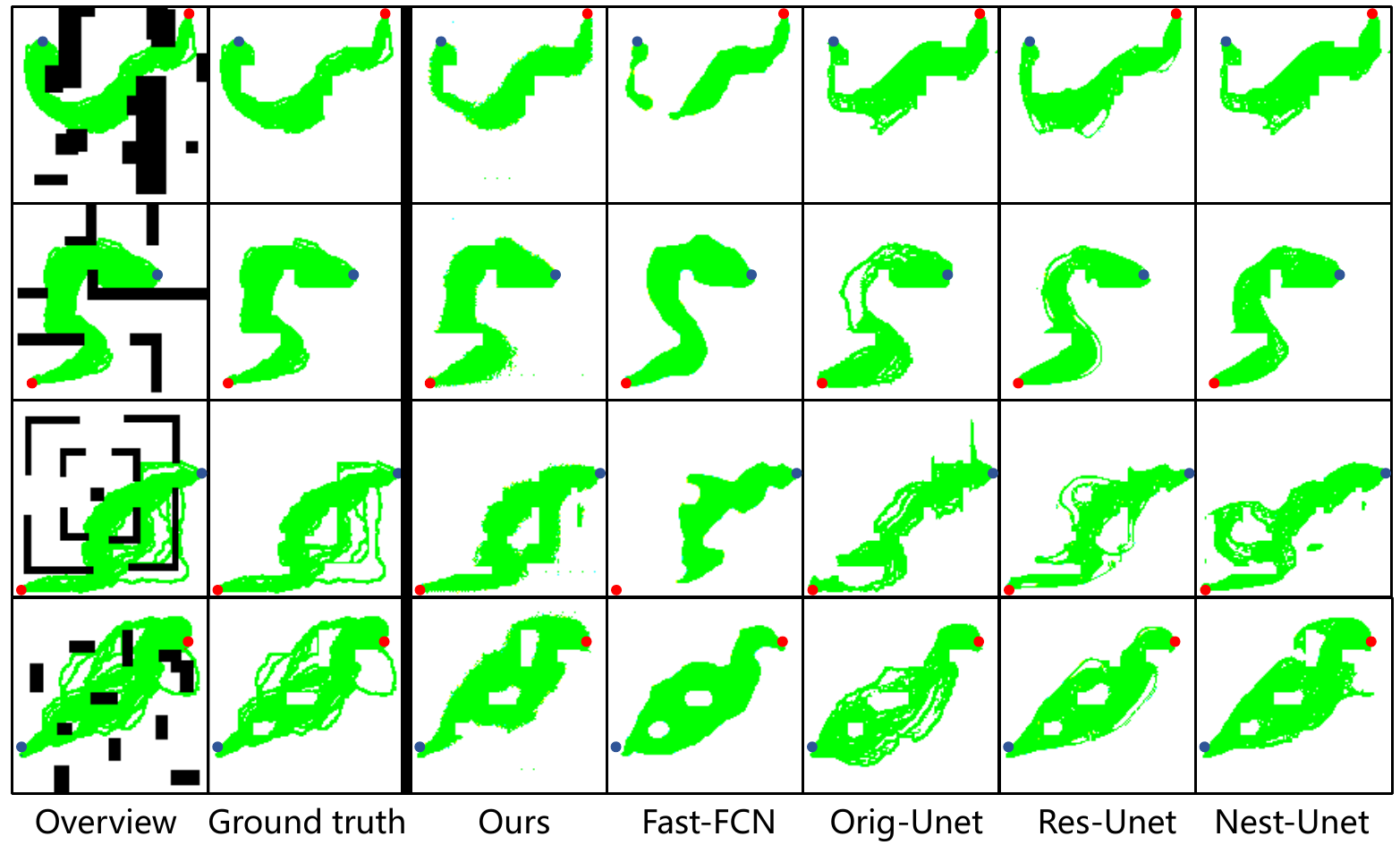}
\caption{The visualized planning areas (i.e., green areas) using our method and other baseline approaches.}
\label{map1}
\end{figure}

Considering the DWConv and the difference in computational logic and data transferring \cite{DSC_cpu}, 
our model achieves significant improvement on the CPU.
We select four typical samples from the test set and display the visual comparison in Fig \ref{map1}, 
which shows our model outperforms others in terms of the connectivity and similarity of planning areas.

\begin{figure}[htbp]
\centering
\includegraphics[width=0.95\columnwidth,height=0.31\linewidth]{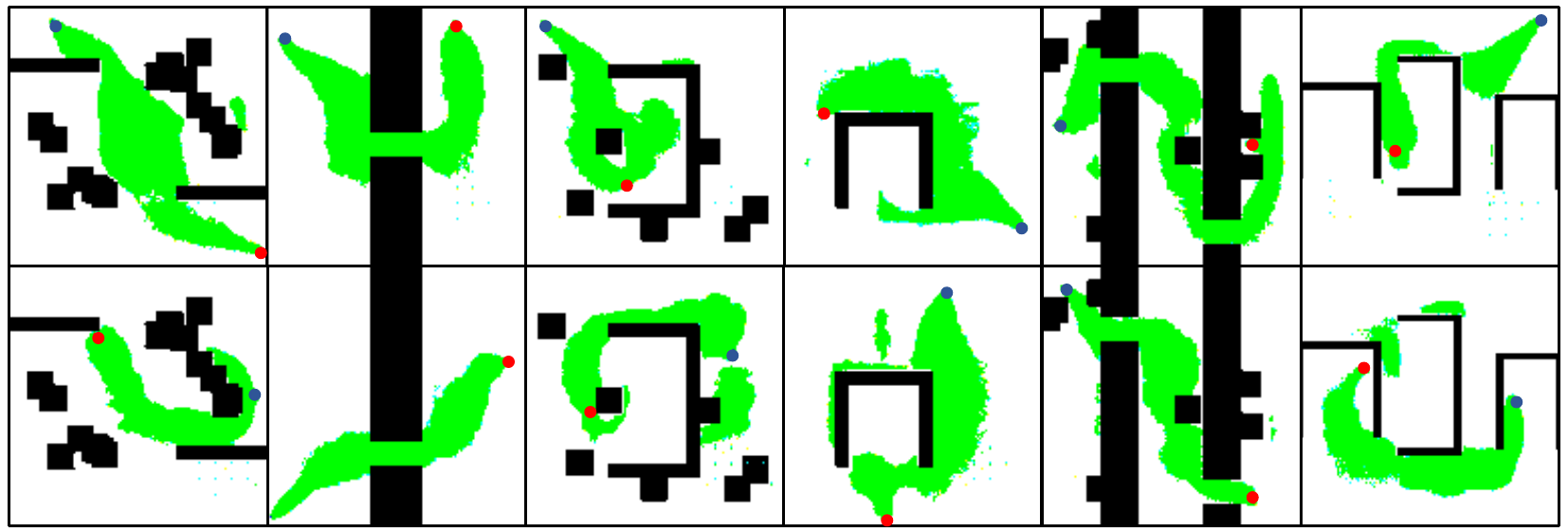}
\caption{The generated guidance maps on the Motion Planing (MP) dataset.}
\label{data2}
\end{figure}

\begin{figure}[htbp]
\centering
\includegraphics[width=0.98\columnwidth,height=0.48\linewidth]{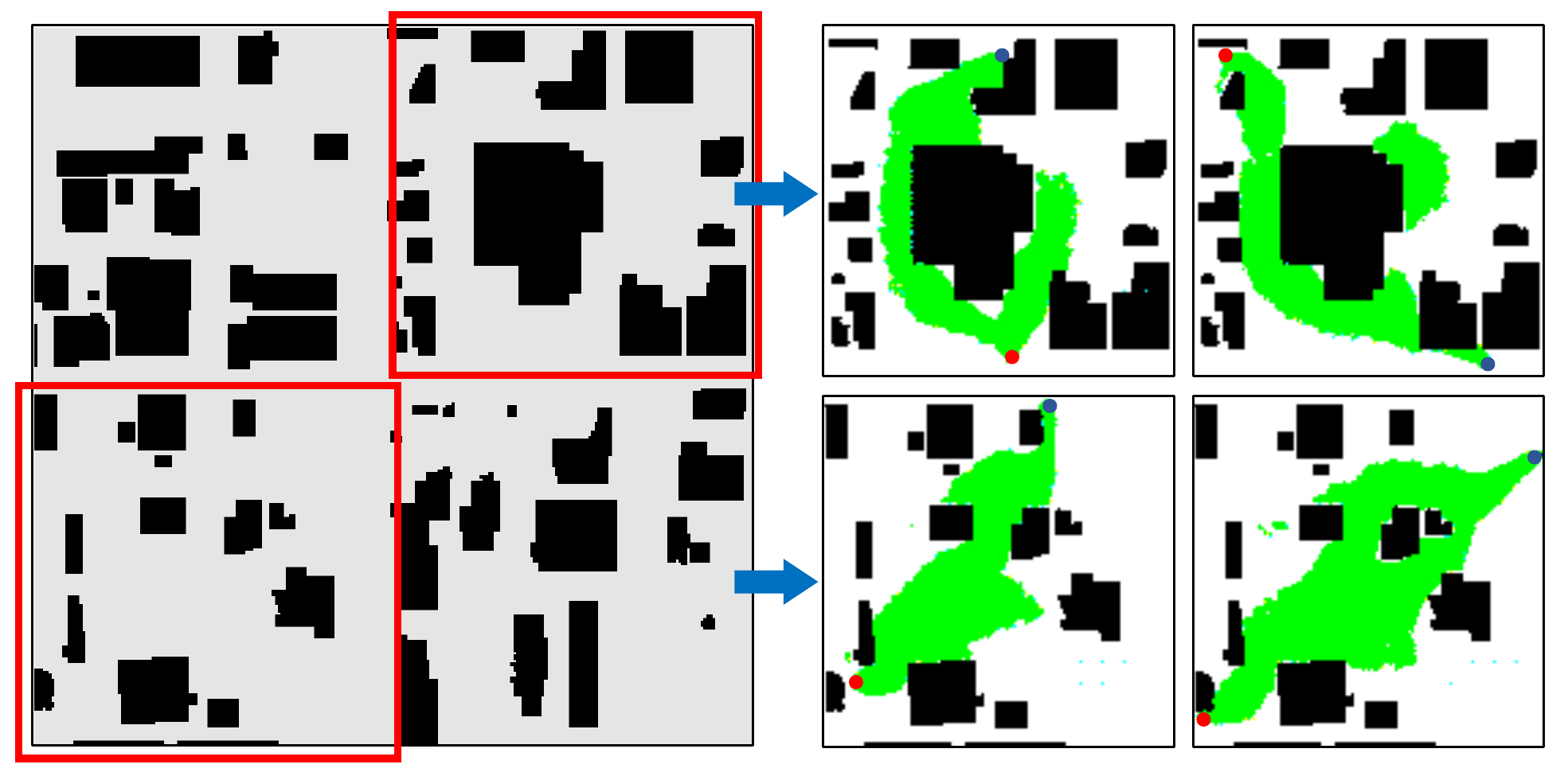}
\caption{The generated guidance maps on the City/Street Map (ASM) Dataset.}
\label{data3}
\end{figure}

\textbf{Generalization on other Dataset:}
To show the generalization capability of our model in new environments, we adopt new maps from two datasets, 
the Motion Planing (MP) dataset\cite{MPData} and City/Street Map (ASM) Dataset\cite{ASMData}, which have never seen before to our model. 
Since the images in ASM have higher resolution, we randomly crop areas from these images as input.
Besides, all maps are resized to $128\times 128$ with a pair of start and end points. Fig. \ref{data2} and Fig. \ref{data3} show that 
our model performs well in new environments by generating connected areas 
between the start point and the end point, which is necessary for the planning of a feasible path.

\textbf{Comparison to classical methods:}
The effectiveness of our method is compared with classical methods, such as RRT, in terms of the path length and number of samples in path planning.
The path length denotes the distance that the robot travels from the start point to the end point, 
and the number of samples is a measure of efficiency 
indicating how many points were sampled from the map until the appropriate path was discovered.

\begin{figure}[htbp]
\begin{minipage}[t]{0.51\linewidth}
    \centering
    \includegraphics[width=\textwidth]{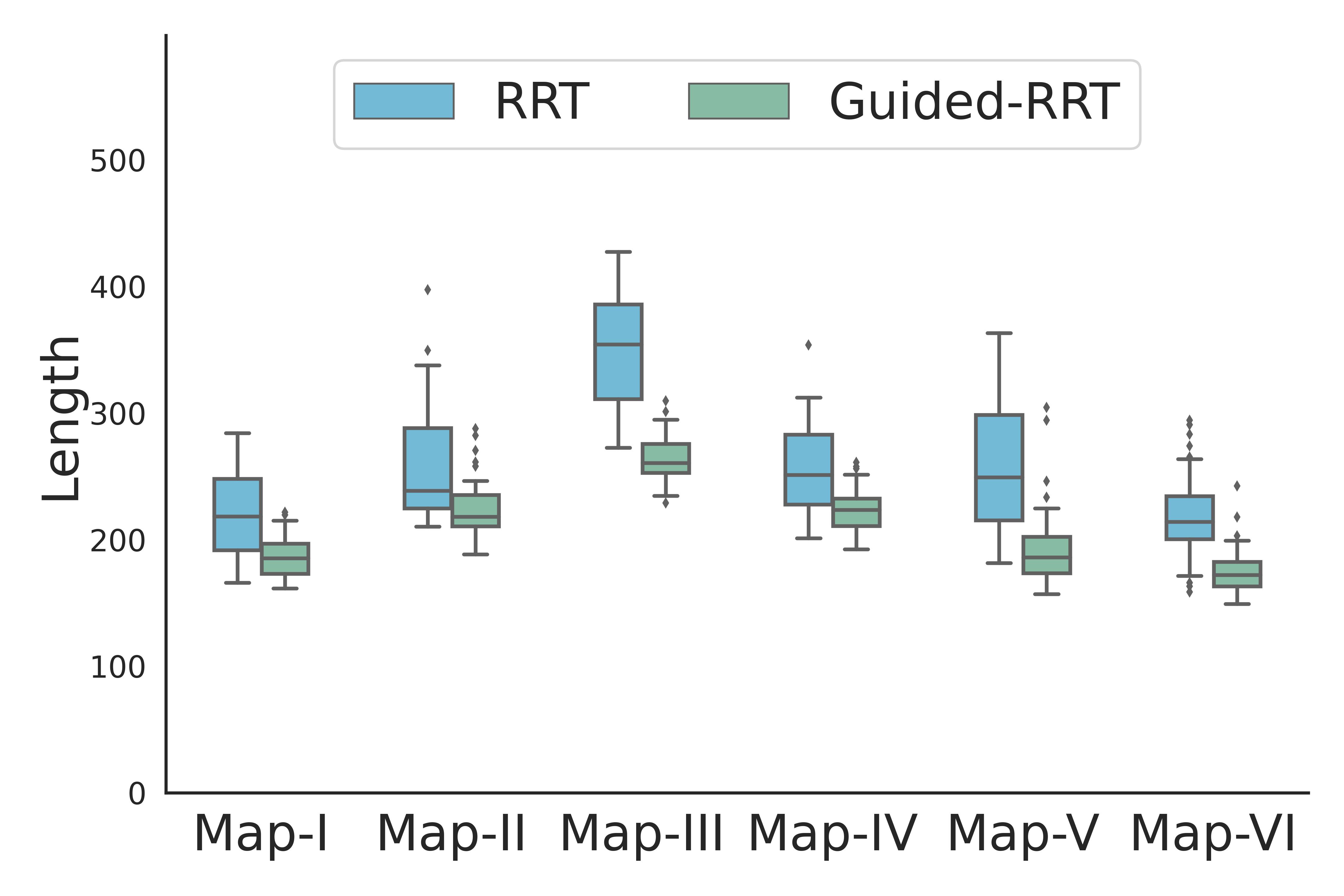}
    \centerline{\footnotesize{(a) Path length}}
\end{minipage}%
\begin{minipage}[t]{0.51\linewidth}
    \centering
    \includegraphics[width=\textwidth]{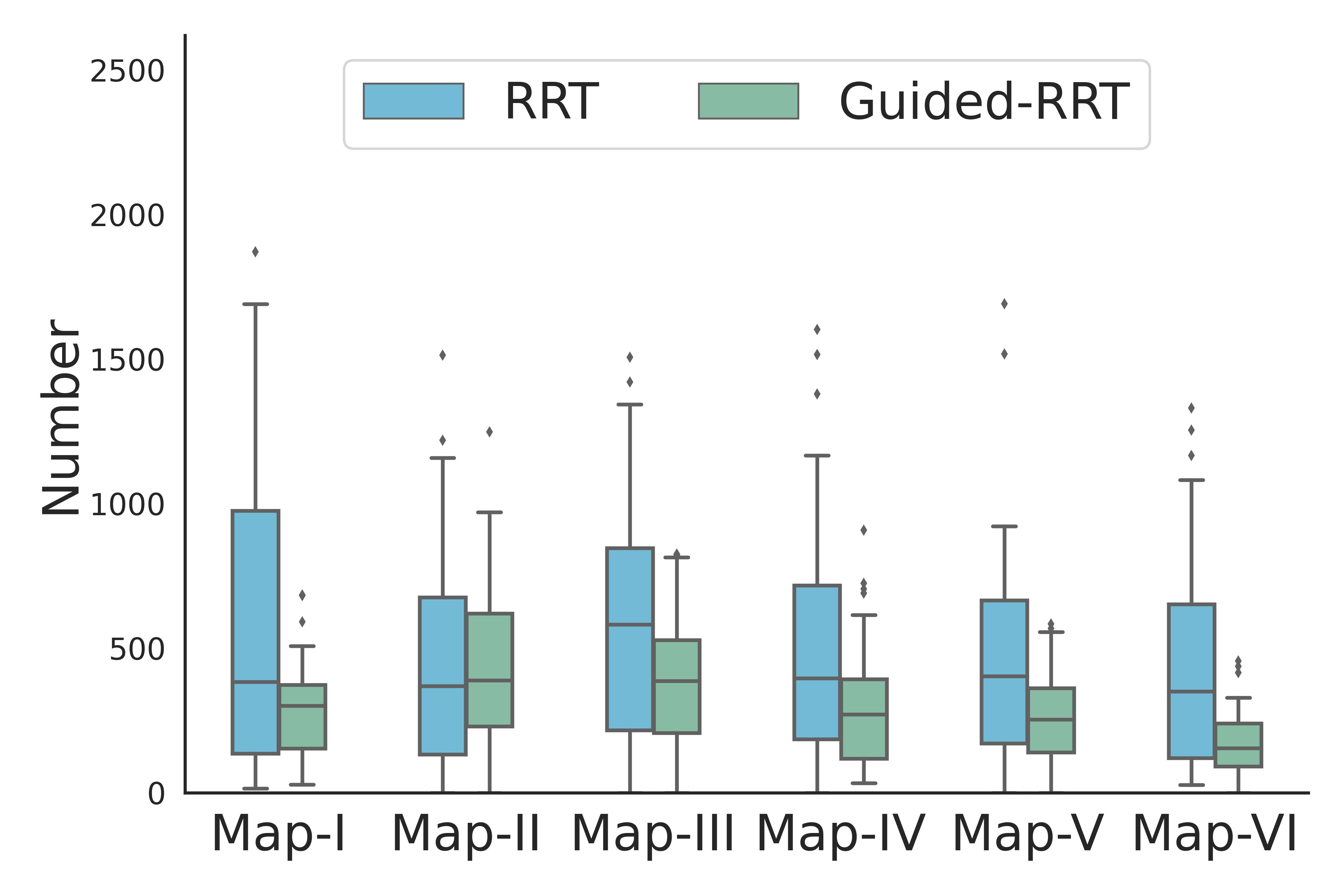}
    \centerline{\footnotesize{(b) Sampled nodes}}
\end{minipage}
\caption{Comparison on path length and sampled nodes}
\label{RRTbox}
\end{figure}

\begin{figure}[htbp]
\centering
\includegraphics[width=0.95\columnwidth,height=0.63\linewidth]{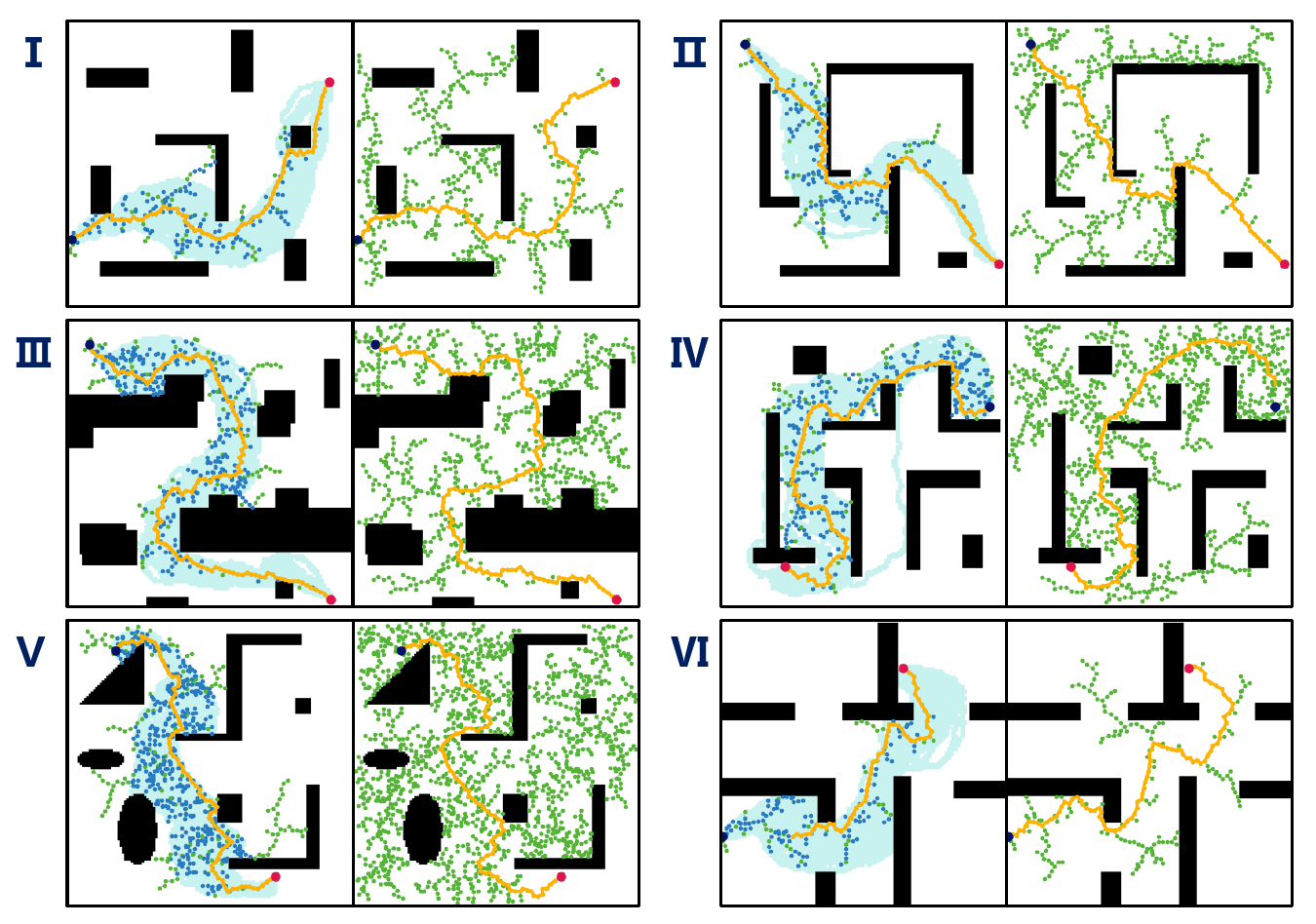}
\caption{Sampling process of $Guided$-$RRT$ (the left column) and $RRT$ (the right column) in planning tasks.}
\label{RRTcpar}
\end{figure}

The searching capabilities of $RRT$ and $Guided$-$RRT$ are evaluated on a set of different task instances, as shown in Fig. \ref{RRTcpar}. 
In particular, both methods have been executed $50$ times with identical step size. 
The statistical results of finding the first path and the path search procedure are displayed in Fig. \ref{RRTbox} and Fig. \ref{RRTcpar}, respectively.
Fig. \ref{RRTbox} indicates that utilizing the designed non-uniform sampling module can encourage path planning in prospective areas where the path is likely to exist and thus save computation and storage resources. Fig. \ref{RRTcpar} shows the sampling process using $RRT$ and $Guided$-$RRT$ on six different types of maps. Clearly, our method is much more efficient in terms of the environment sampling.

\begin{figure}[htbp]
\begin{minipage}[t]{0.51\linewidth}
    \centering
    \includegraphics[width=\textwidth]{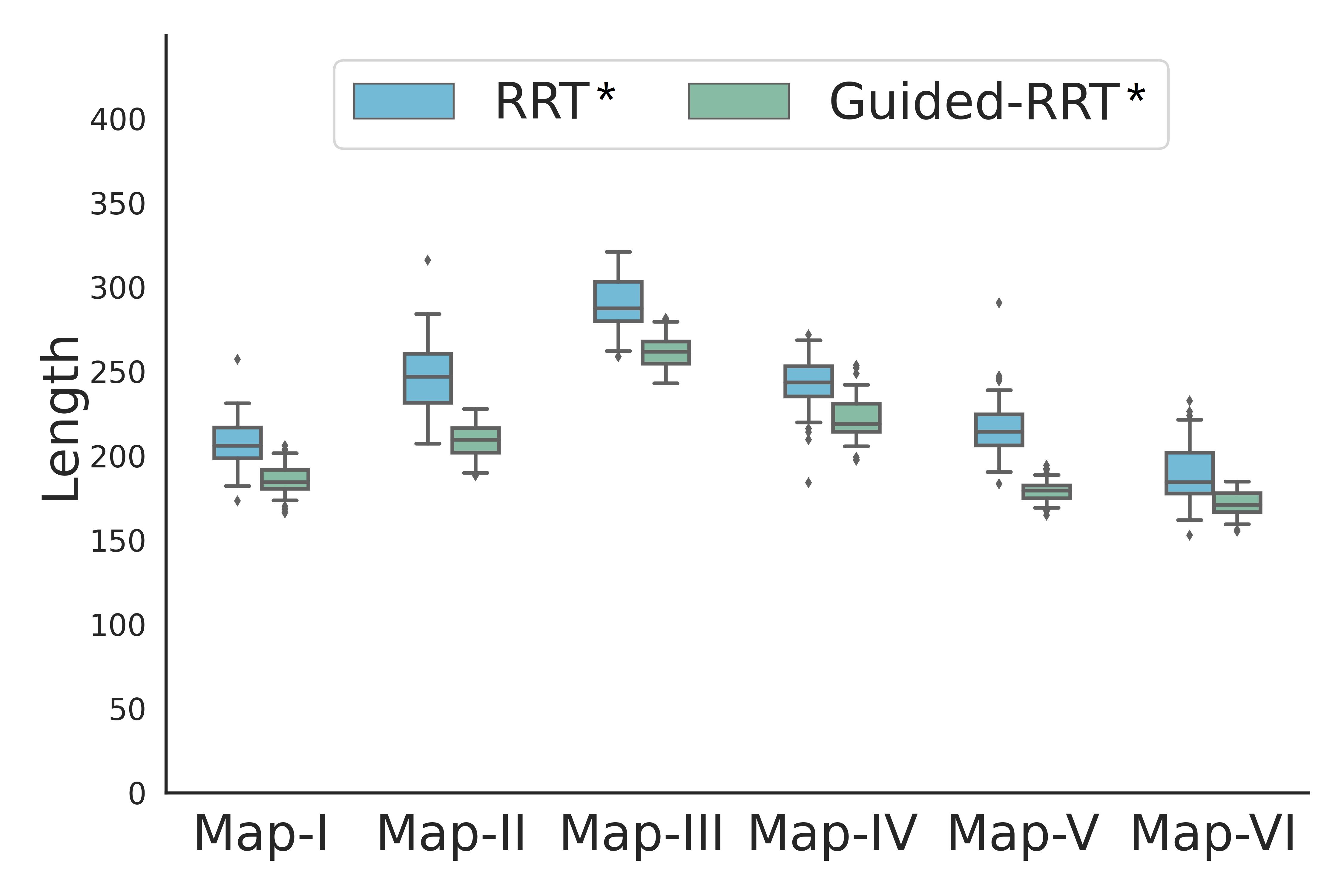}
    \centerline{\footnotesize{(a) Path length}}
\end{minipage}%
\begin{minipage}[t]{0.51\linewidth}
    \centering
    \includegraphics[width=\textwidth]{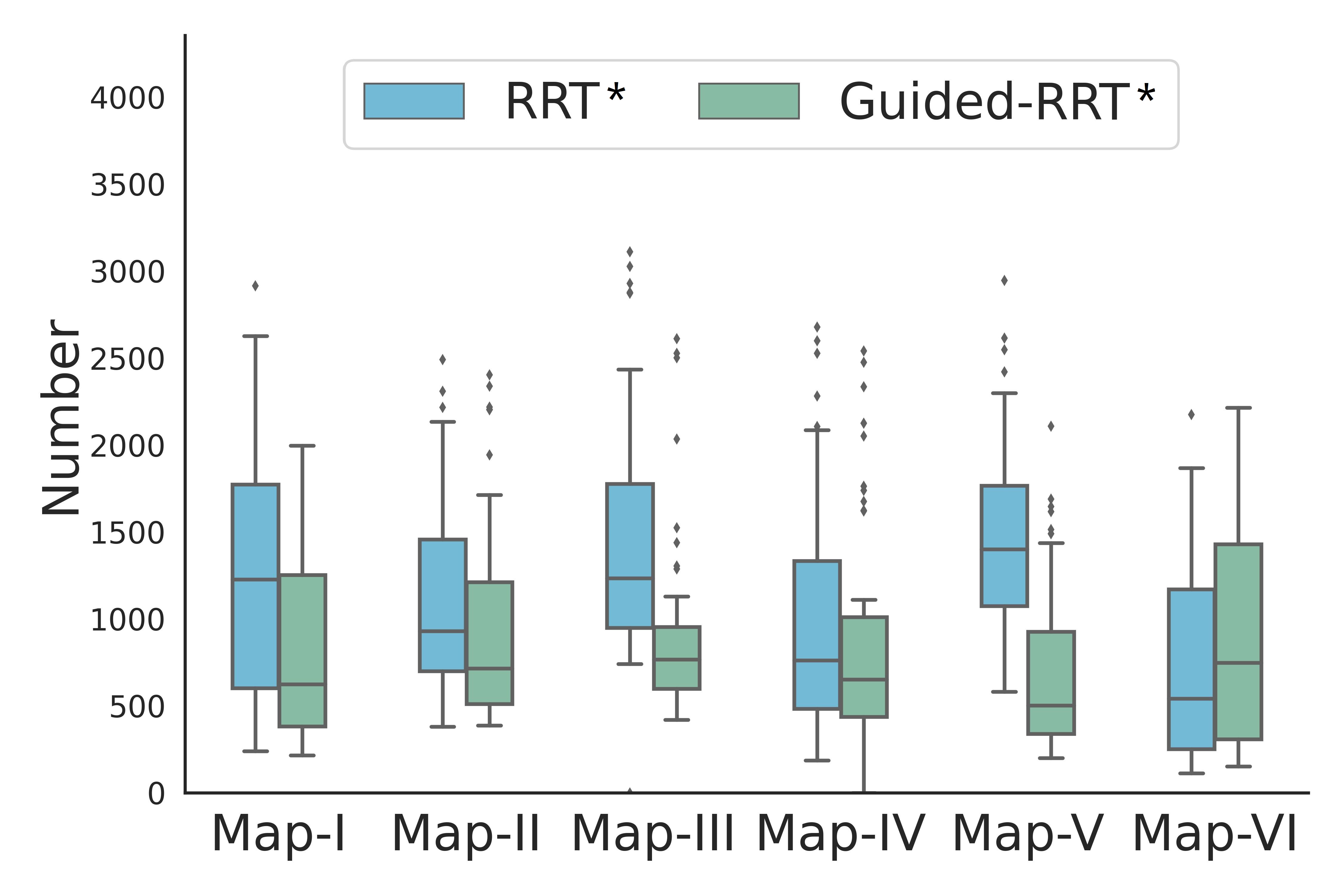}
    \centerline{\footnotesize{(b) Sample nodes}}
\end{minipage}
\caption{Comparison on optimal path length and sample nodes}
\label{RRTSbox}
\end{figure}

\begin{figure}[htbp]
\centering
\includegraphics[width=0.95\columnwidth,height=0.63\linewidth]{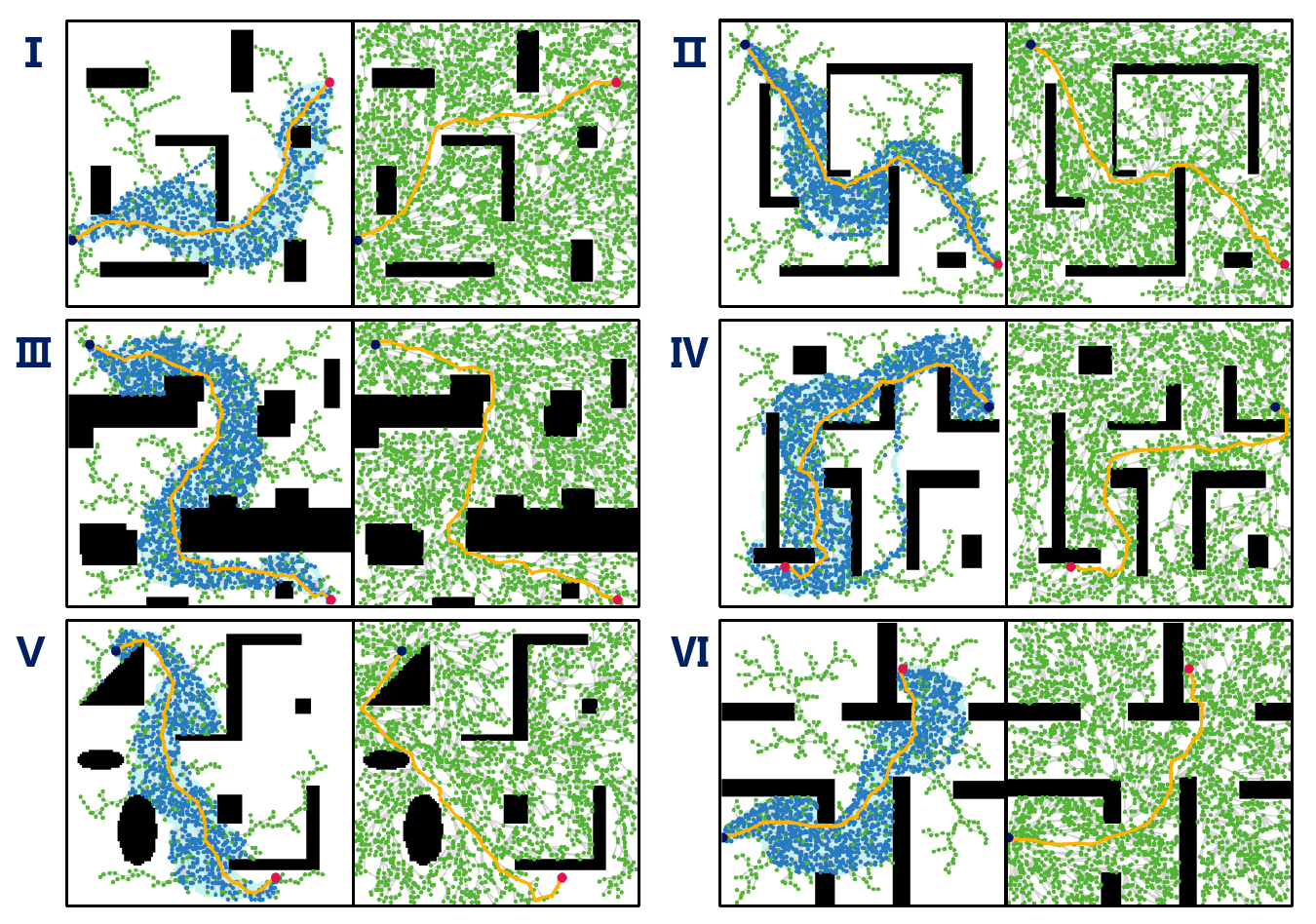}
\caption{Sampling process of $Guided$-$RRT^{*}$ (the left column) and $RRT^{*}$ (the right column) in optimal planning tasks.}
\label{RRTScpar}
\end{figure}

Similar comparisons are also performed for $RRT^{*}$ and $Guided$-$RRT^{*}$. 
The results are listed in Fig. \ref{RRTSbox}.
Unlike $RRT$ and $Guided$-$RRT$, 
it is more concerned with finding the optimal path in a finite iterations. Fig. \ref{RRTSbox} indicates that our method outperforms $RRT^{*}$ in the sense we can find shorter paths but with less sampled nodes. Fig. \ref{RRTScpar} shows the sampling process using $RRT^{*}$ and $Guided$-$RRT^{*}$ on a set of different types of maps. Without sampling the entire map, $Guided$-$RRT^{*}$ can converge faster to the optimal path with fewer sampled nodes. This is more obvious in the complex maps such as map-\uppercase\expandafter{\romannumeral3} and map-\uppercase\expandafter{\romannumeral5}. In these scenarios, since the robot needs to traverse a longer space, using the guidance map to steer the search will significantly improve the planning. In short, conventional approaches, such as $RRT$ and $RRT^{*}$, search the entire map, while our method boosts the search by only focusing on prospective regions where a feasible planning might exist.

\subsection{Real-world Demonstrations}
\begin{figure}[htbp]
\centering
\includegraphics[width=1\columnwidth,height=0.32\linewidth]{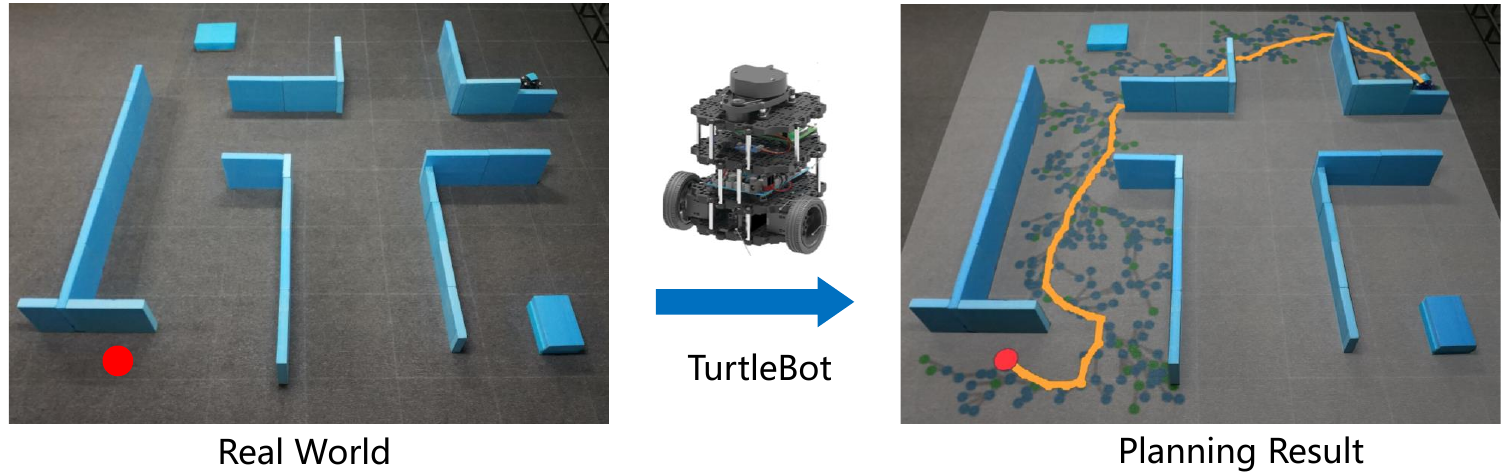}
\caption{Path planning in a physical environment.}
\label{realw}
\end{figure}

To further demonstrate the path planning capability of our approach in real-world settings, the trained model is deployed on the TurtleBot platform, running Ubuntu system. The turtlebot is localized using the embedded IMU.
The start and end points are assigned randomly, and the guidance map is generated through the network. 
After applying the $Guided$-$RRT*$ algorithm, the generated trajectory is shown in  
Fig.\ref{realw}. More details can be found in the experiment video\footnote{Experiment video is available online at: \url{https://youtu.be/pHdR1BHOLcA}}.


\section{Conclusion}
This work presents a learning-based path planning algorithm for resource-constrained mobile robots, which involves a lightweight deep neural network and a bias-sampling planner. To improve the search efficiency, the network learns the guidance map from the demonstration, so that nodes can be sampled within prospective areas rather than the entire space. The use of ShuffleNet-units and deeply separable convolutions enables the network to achieve fast inference and better generation of prospective regions while reducing the complexity and number of parameters, making it possible to be deployed on resource-constrained robot platforms. In addition, we construct a publicly available path-planning dataset with successful experience to foster the development of learning-based planning methods. Future research will consider extending the current approach by integrating temporal logic specifications to address path planning for more challenging tasks beyond point-to-point navigation.

\bibliography{ref}

\end{document}